\documentclass[10pt,twocolumn,letterpaper]{article}

 \usepackage[pagenumbers]{cvpr} 

\usepackage{tabularx}

\usepackage{booktabs}
\usepackage{colortbl}

\definecolor{cvprblue}{rgb}{0.21,0.49,0.74}
\usepackage[pagebackref,breaklinks,colorlinks,allcolors=cvprblue]{hyperref}

\def\confName{CVPR}
\def\confYear{2027}

\title{CamAct: Learning Robotic Manipulation from In-the-Wild Videos  \\ via Camera Motion}

\author{
    Haoming Gu$^{1,3}$ \quad
    Yunpeng Qing$^{2,3}$ \quad
    Zhi Hou$^{3\dagger}$ \quad
    Weinan Zhang$^{1\dagger}$ \\[0.4em]
    $^{1}$Harbin Institute of Technology \quad
    $^{2}$Zhejiang University \quad
    $^{3}$ACE Robotics \\[0.2em]
}

\begin{document}
\maketitle

\begingroup
\renewcommand{\thefootnote}{\fnsymbol{footnote}}
\footnotetext[2]{Corresponding authors.}
\endgroup

\begin{abstract}
Vision-Language-Action (VLA) models benefit from large, diverse datasets for robotic manipulation, yet collecting robot demonstrations remains costly. Recent work has explored egocentric videos as a source of supervision, primarily drawing on demonstrations of human–object interaction. We show that in-the-wild videos can improve robotic manipulation even when they contain no demonstrations of object manipulation. Specifically, camera motion recovered from these videos provides an additional source of motion supervision that complements robot demonstrations. Building on this finding, we introduce \textbf{CamAct}, which incorporates this camera-motion supervision into VLA policies alongside robot demonstrations, extending the use of human videos for manipulation learning beyond object interactions. Experiments show that incorporating this video supervision during fine-tuning improves success rates from 96.40\% to 98.10\% on LIBERO and from 22.96\% to 29.32\% on RoboCasa 365. The resulting policy also demonstrates strong results in out-of-distribution generalization on LIBERO-Plus and real-world manipulation tasks.
\end{abstract}

\section{Introduction}
\label{sec:intro}

\begin{figure}[t]
    \centering
    \includegraphics[width=\linewidth]{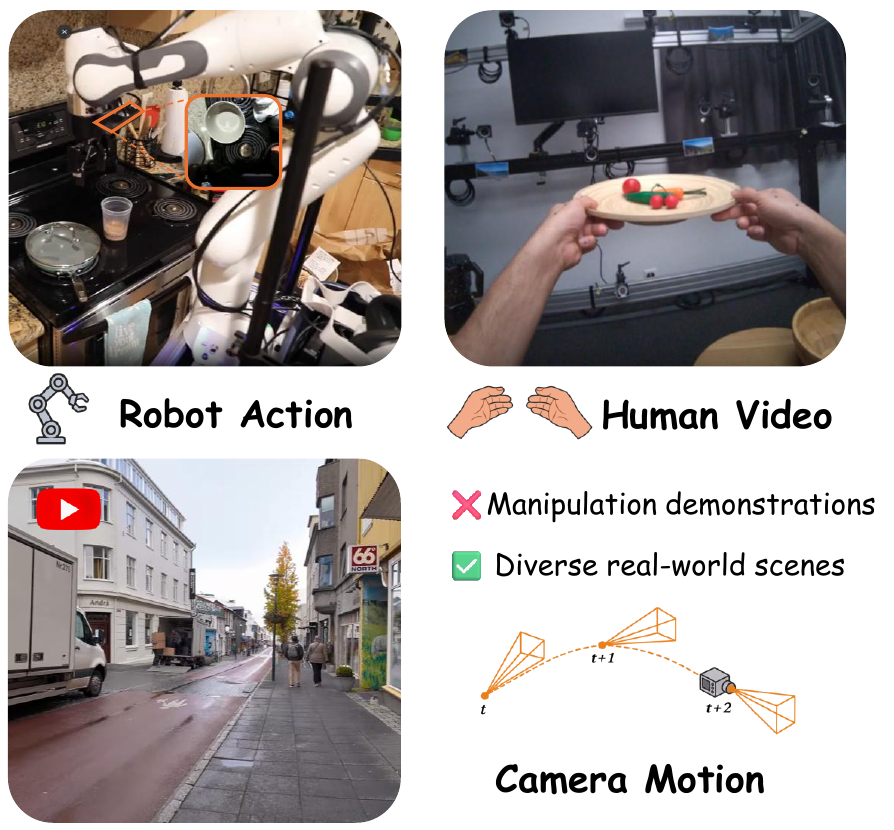}
\caption{\textbf{Beyond manipulation demonstrations.}
Robot action and human video
provide demonstrations of manipulation.
In-the-wild videos capture diverse real-world
scenes without manipulation demonstrations.
CamAct uses camera motion recovered from these videos
as additional supervision for robotic manipulation.}
    \label{fig:fig1}
\end{figure}

Developing manipulation policies that generalize across tasks and environments is a central goal in robotics. Vision-language-action (VLA) models unify visual perception, language understanding, and action generation within a single policy for robotic manipulation~\cite{kim2024openvla,black2024pi0}. These models benefit from diverse training data for generalization, including robot demonstrations collected in real-world environments, as well as egocentric human videos~\cite{bjorck2025gr00t,li2026ace}. Although recent efforts have built large-scale datasets from demonstrations collected across robot embodiments~\cite{o2024open,khazatsky2024droid}, scaling robot demonstration data remains challenging due to the high cost of real-world collection and annotation, as well as embodiment-specific requirements. Egocentric human videos provide a lower-cost source of supervision for robot learning, yet existing approaches primarily focus on human--object interactions to complement robot demonstrations~\cite{kareer2025egomimic,punamiya2025egobridge}. In contrast, large language models benefit from diverse web-scale data~\cite{bai2025qwen3, gpt, opus}, motivating us to explore whether robotic manipulation can benefit from broader video sources, including egocentric videos without explicit object manipulation demonstrations (Fig.~\ref{fig:fig1}).

Web-scale data have been widely used to train vision-language models (VLMs), with recent efforts further incorporating such data into VLA pretraining or adapting VLMs for downstream policy learning through vision-language tasks~\cite{intelligence2025pi_,yang2026vlaser,du2026embodiedmidtrain}. However, generalization in robotic manipulation requires understanding not only visual semantics but also the motion and dynamics of environments. This motivates us to explore whether the rich motion information in web videos can further benefit manipulation policy learning. CityWalker~\cite{liu2025citywalker} demonstrates that camera trajectories recovered from in-the-wild videos can provide effective motion supervision for navigation policies. Moreover, recent unified frameworks show that jointly training on navigation and manipulation trajectories can benefit both domains~\cite{yang2024pushing,wang2026qwen,zhang2026onevla}. Building on these insights, we investigate whether camera motion from in-the-wild walking videos can provide additional supervision for manipulation policies, even without explicit object interactions.

As humans walk through real-world environments, egocentric cameras capture rich spatial and motion information through changing viewpoints, as wrist-mounted cameras do during robotic manipulation. Although these videos contain no explicit object interactions, their camera motion provides a potential source of supervision for learning transferable spatial and motion representations. Building on this observation, we propose \textbf{CamAct}, a unified VLA framework that jointly learns from robot demonstrations and camera motion in in-the-wild videos. Specifically, we recover camera trajectories offline and convert them into pseudo-actions, allowing both robot actions and camera-derived pseudo-actions to be learned through a shared VLA backbone and action head. Through this joint training, CamAct aims to improve manipulation generalization by learning from diverse real-world camera motion.

We evaluate CamAct on LIBERO~\cite{liu2023libero} and RoboCasa 365~\cite{nasiriany2026robocasa365} using StarVLA~\cite{community2026starvla} as the base model. Joint training with camera motion improves success rates from 96.40\% to 98.10\% on LIBERO and from 22.96\% to 29.32\% on RoboCasa 365. CamAct further demonstrates improved out-of-distribution generalization on LIBERO-Plus~\cite{fei2025libero} and promising performance on real-world manipulation tasks. Extensive ablations and motion distribution analyses examine how video characteristics and training configurations influence downstream performance.

Our contributions are summarized as follows:

\begin{itemize}
    \item We explore the potential of in-the-wild videos for robotic manipulation beyond visual-semantic learning, showing that camera motion can provide useful supervision even without explicit object interactions.

    \item We propose \textbf{CamAct}, a unified VLA framework that jointly learns from robot demonstrations and camera-derived pseudo-actions, demonstrating consistent improvements in manipulation performance and out-of-distribution generalization.

    \item We conduct extensive ablations and motion distribution analyses to investigate how video characteristics, data composition, and training design choices affect downstream manipulation performance.
\end{itemize}




\section{Related Work}
\label{sec:related_work}

\subsection{Vision-Language-Action Models}

Robot learning has progressed from task-specific imitation policies toward generalist robot policies and
vision-language-action (VLA) models trained on large and diverse datasets. RT-1~\cite{Brohan-RSS-23} demonstrated scalable transformer-based control from real-robot demonstrations, while RT-2~\cite{brohan2023rt} connected web-scale vision-language knowledge with robot action prediction. Open X-Embodiment and RT-X~\cite{o2024open} further brought together trajectories from different robot platforms and demonstrated the benefits of cross-embodiment training. Subsequent systems, including Octo, OpenVLA, $\pi_0$, RDT-1B, and CogACT~\cite{team2024octo,kim2024openvla,black2024pi0,liu2024rdt1b,li2024cogact}, as well as GR00T, Qwen-VLA, and MolmoAct2~\cite{bjorck2025gr00t,wang2026qwen,fang2026molmoact2}, have further developed scalable policy architectures and action-generation mechanisms for generalist manipulation. Related efforts improve action representations and policy adaptation~\cite{pertsch2025fast,kim2025oft}.

Alongside these developments, joint training enables VLA models to learn from heterogeneous data sources and multiple learning objectives. Lin et al.~\cite{lin2026systematic} systematically examine how data modalities and training strategies affect joint training for robotic manipulation. Recent models jointly learn from demonstrations collected across robot embodiments~\cite{zheng2026x,team2025gemini}. Beyond combining robot datasets, $\pi_{0.5}$ integrates robot action learning with vision-language tasks and high-level semantic prediction~\cite{intelligence2025pi_}. Auxiliary tasks further extend this paradigm to visual grounding and spatial reasoning~\cite{zawalski2025ecot,chen2025internvla,team2026ace}, as well as 3D geometric supervision~\cite{peng2026g3vla,shi2026threedthinkvla}. Other approaches jointly train action prediction with future visual prediction~\cite{wu2026foundation,cai2026internvla,li2026dypesvla} or 3D spatial trajectory prediction~\cite{kim2026pri4r,he2026elan4d}.

\subsection{Learning Robotic Manipulation from Videos}

Human videos provide a scalable source of manipulation experience beyond robot-collected demonstrations. Earlier approaches use internet manipulation videos and human play to guide robot policy learning~\cite{shaw2023videodex,wang2023mimicplay}. Recent methods jointly learn from human and robot demonstrations while addressing cross-domain alignment~\cite{kareer2025egomimic,punamiya2025egobridge}, with further efforts studying the scaling and integration of egocentric supervision~\cite{zheng2026egoscale,wu2026atomego}. Camera-space action representations also facilitate human--robot transfer~\cite{li2026ace,pan2026ironmind}. Other approaches construct robot-aligned observations and action supervision through motion reconstruction, retargeting, and visual synthesis~\cite{video,jeong2026huro,wang2026ego2robot,guo2026roboedit}. VidAct~\cite{li2026vidact} reconstructs object motion to generate simulated robot demonstrations.

These approaches primarily derive supervision from demonstrated object interactions, leaving motion in videos without manipulation demonstrations less explored for manipulation learning. Camera trajectories provide another form of motion supervision: CityWalker and Kuno et al. recover camera motion from walking videos for robot navigation~\cite{liu2025citywalker,kuno2026cotraining}. Within manipulation, ActiveMimic~\cite{lin2026activemimic} jointly predicts camera and wrist trajectories from manipulation videos to learn active perception and manipulation. In contrast, CamAct jointly learns from robot demonstrations and camera-derived pseudo-actions from in-the-wild walking videos that contain no manipulation demonstrations.
\section{Methodology}
\label{sec:method}

\begin{figure*}[t]
    \centering
    \includegraphics[width=\linewidth]{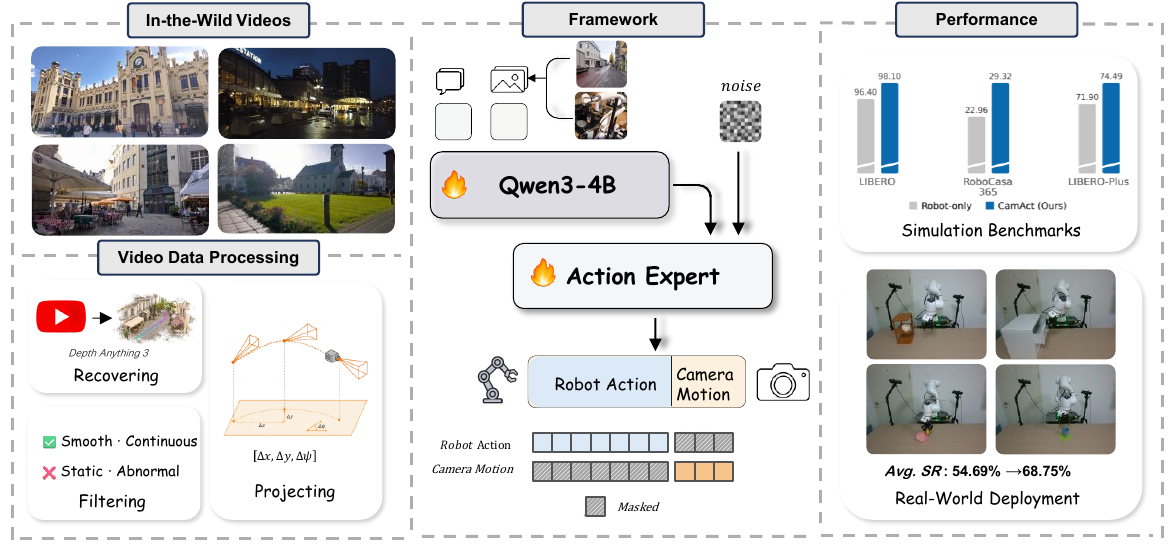}
    \caption{CamAct converts camera trajectories from in-the-wild videos into pseudo-actions for joint training with robot demonstrations, improving manipulation performance on simulation benchmarks and real-world tasks.}
    \label{fig:fig2}
\end{figure*}

CamAct is a unified VLA framework that jointly learns from robot demonstrations and camera motion in in-the-wild videos (Fig.~\ref{fig:fig2}). We recover camera trajectories offline and convert them into pseudo-actions that provide additional motion supervision for policy learning.

\subsection{Problem Formulation}
\label{sec:formulation}

We consider a VLA model, where a policy predicts future actions from visual observations and a task instruction. At time step $t$, the visual observations $\mathbf{o}_t$ consist of images from the available camera views, and $\ell$ denotes the task instruction. The policy $\pi_{\theta}(\mathbf{A}_t \mid \mathbf{o}_t,\ell)$ models a distribution over action chunks
\begin{equation}
    \mathbf{A}_t =
    [\mathbf{a}_t,\ldots,\mathbf{a}_{t+H-1}]
    \in \mathbb{R}^{H\times d_R},
\end{equation}
where $H$ is the prediction horizon and
$d_R$ is the robot action dimension.
We learn this policy from robot demonstrations
$\mathcal{D}_R=\{(\mathbf{o}_t,\ell,\mathbf{A}_t)\}$.

In addition to robot demonstrations, we use in-the-wild videos without manipulation demonstrations. Camera trajectories recovered from these videos provide pseudo-actions, yielding a dataset
$\mathcal{D}_V=\{(I_t,\mathbf{U}_t)\}$. Here, $I_t$ is the current video frame and
$\mathbf{U}_t=[\mathbf{u}_t,\ldots,\mathbf{u}_{t+H-1}]$
contains the subsequent $H$ camera-motion pseudo-actions, whose construction is described in Sec.~\ref{sec:video_data}. Our goal is to improve the manipulation policy by jointly learning from robot actions in $\mathcal{D}_R$ and camera-motion pseudo-actions in $\mathcal{D}_V$.

\subsection{Video Data Processing}
\label{sec:video_data}

\paragraph{Video Data.}
To investigate whether camera motion can provide additional supervision for robotic manipulation, we use first-person walking videos sourced from the internet. Such videos are widely available online and capture people moving through diverse real-world environments. As the camera moves with the person, these recordings capture translations, turns, and changes in viewpoint without manipulation demonstrations. Motivated by their availability and the diversity of scenes and motions they capture, we assemble a collection totaling 44.40 hours. After filtering, 36.09 hours of valid video remain.

\paragraph{Recovering Camera Motion.}
To obtain camera-motion supervision from the collected
videos, we recover camera trajectories using
Depth Anything~3~\cite{lin2025depth}.
We address two practical issues during this process.
First, the videos may contain abrupt visual transitions
and sustained near-static intervals.
We detect both using pixel differences between consecutive
frames, partition each video into visually continuous shots,
and exclude near-static footage before pose estimation.
Each shot is then processed independently.
Second, long shots are reconstructed over overlapping
temporal windows, whose reconstructions may differ in
scale, rotation, and translation.
We estimate Sim(3) transformations from their shared frames
to align adjacent reconstructions and assemble a single
trajectory for each shot.
We denote the recovered camera-to-world poses by
$\mathbf{T}_t=(\mathbf{R}_t,\mathbf{c}_t)\in\mathrm{SE}(3)$,
where $\mathbf{R}_t$ and $\mathbf{c}_t$ represent the
camera orientation and position in the shot's
reconstruction coordinate frame.

\paragraph{Constructing Pseudo-Actions.}
The collected walking videos primarily capture movement
along the ground plane and changes in heading, with
relatively small vertical oscillations and variations
in camera tilt.
We therefore represent the recovered camera motion using
planar translation and yaw rotation in $\mathrm{SE}(2)$.
For each shot, we project camera positions and forward
directions onto a fixed plane whose normal is the initial
camera-up direction, assuming that the camera is approximately
level in the first frame.

After trajectory filtering and smoothing, let
$\mathbf{p}_t\in\mathbb{R}^{2}$ and $\psi_t$ denote the
planar position and unwrapped heading.
We define each pseudo-action as the relative motion
between consecutive time steps:
\begin{equation}
    \mathbf{u}_t =
    \begin{bmatrix}
        R_{\mathrm{2D}}(\psi_t)^{\top}
        (\mathbf{p}_{t+1}-\mathbf{p}_t) \\
        \psi_{t+1}-\psi_t
    \end{bmatrix}
    \in\mathbb{R}^{3},
    \label{eq:planar_increment}
\end{equation}
where $R_{\mathrm{2D}}(\psi_t)$ maps the camera's
heading-aligned forward-left axes to the fixed planar
reference frame.
The three components correspond to forward displacement,
leftward displacement, and heading change.
Each $\mathbf{u}_{t+k}$ uses the camera heading at time
$t+k$, so a chunk describes successive local motion
increments.
These increments are independent of the shot's planar
origin and reference orientation.

We pair each frame $I_t$ with the subsequent $H$ pseudo-actions
to construct $\mathbf{U}_t=[\mathbf{u}_t,\ldots,\mathbf{u}_{t+H-1}]$,
requiring the complete target chunk to remain within
a single valid segment.
Details of temporal sampling, trajectory filtering, and
smoothing are provided in the supplementary material.

\subsection{Policy Learning with Camera Motion}
\label{sec:policy}

\paragraph{Policy Architecture.}
We build CamAct on StarVLA~\cite{community2026starvla},
using a Qwen3-VL-4B vision-language backbone~\cite{bai2025qwen3}
and a flow-matching action head.
Both components are shared by robot demonstrations and
video samples.
The backbone encodes each input into conditioning features
$\mathbf{h}_t=f_{\theta}(\mathbf{x}_t)$,
where $\mathbf{x}_t$ denotes either a robot input
$(\mathbf{o}_t,\ell)$ or a video input
$(I_t,\varnothing)$, with $\varnothing$ denoting an empty
instruction.
Conditioned on $\mathbf{h}_t$, the shared action head
generates action chunks through conditional flow matching.
Neither source uses proprioceptive inputs or explicit
source embeddings.

\paragraph{Action Representation.}
Although we retain planar translation and yaw rotation
in Sec.~\ref{sec:video_data}, camera-derived pseudo-actions
differ from robot end-effector (EEF) actions in motion
magnitude and distribution.
Separate normalization adjusts their numerical ranges
but does not establish a coordinate-wise correspondence
between the two action spaces.
Directly mapping camera-derived pseudo-actions to the
robot's $x$, $y$, and yaw action dimensions may therefore
introduce mismatched supervision.
We concatenate the two representations, preserving
separate output dimensions while sharing a single action head.

Specifically, we normalize robot actions and camera-derived
pseudo-actions separately using the 1st and 99th percentiles
of their respective data distributions.
Let $\tilde{\mathbf{a}}_t$ and $\tilde{\mathbf{u}}_t$
denote the normalized robot action and camera-derived
pseudo-action, respectively.
We construct the per-step targets as
\begin{equation}
    \mathbf{y}^{R}_t =
    [\tilde{\mathbf{a}}_t;\mathbf{0}_{3}],
    \qquad
    \mathbf{y}^{V}_t =
    [\mathbf{0}_{d_R};\tilde{\mathbf{u}}_t],
    \label{eq:packed_targets}
\end{equation}
where the semicolon denotes concatenation.
A single shared action head, including its final output
projection, predicts chunks in
$\mathbb{R}^{H\times(d_R+3)}$.
A loss mask restricts supervision to the first $d_R$
dimensions for robot samples and the last three dimensions
for video samples.
An ablation comparing direct coordinate alignment with
concatenation is reported in the supplementary material.

\begin{table*}[t]
\centering
\caption{Evaluation results on the LIBERO and RoboCasa 365
benchmarks. We report success rates (\%) over 50 rollouts
per task.}
\label{tab:manipulation_success}
\small
\setlength{\tabcolsep}{5pt}

\begin{tabular}{lccccccccc}

\toprule
& \multicolumn{5}{c}{\textbf{LIBERO}}
& \multicolumn{4}{c}{\textbf{RoboCasa 365}} \\
\cmidrule(lr){2-6}
\cmidrule(l){7-10}

Method
& Spatial & Object & Goal & Long & Overall
& Atomic-S
& Composite-S
& Composite-U
& Overall \\

\midrule

Diffusion Policy~\cite{chi2025diffusion}
& 78.30 & 92.50 & 68.30 & 50.50 & 72.40
& 15.70 & 0.20 & 1.30 & 6.10 \\

$\pi_0$~\cite{black2024pi0}
& 96.80 & 98.80 & 95.80 & 85.20 & 94.15
& 34.60 & 6.10 & 1.10 & 14.80 \\

$\pi_{0.5}$~\cite{intelligence2025pi_}
& 98.80 & 98.20 & 98.00 & 92.40 & 96.85
& 39.60 & 7.10 & 1.20 & 16.90 \\

GR00T N1.6~\cite{bjorck2025gr00t}
& 97.70 & 98.50 & 97.50 & 94.40 & 97.00
& 51.10 & 9.40 & 1.70 & 21.90 \\

StarVLA-$\pi$~\cite{community2026starvla}
& 97.60 & 99.80 & 96.80 & 93.40 & 96.90
& 54.67 & 11.63 & 5.50 & 25.16 \\

StarVLA-GR00T~\cite{community2026starvla}
& 99.40 & 99.60 & 95.40 & 91.20 & 96.40
& 52.89 & 9.00 & 3.25 & 22.96 \\

\midrule

\rowcolor[gray]{0.93}
CamAct ($\pi$-base, Ours)
& 99.00 & 99.60 & 97.60 & 96.20 & \textbf{98.10}
& 59.00 & 10.75 & 9.38 & 27.68 \\

\rowcolor[gray]{0.93}
CamAct (GR00T-base, Ours)
& 98.80 & 99.40 & 97.40 & 96.80 & \textbf{98.10}
& 60.11 & 17.00 & 7.00 & \textbf{29.32} \\

\bottomrule
\end{tabular}
\end{table*}

\paragraph{Joint Training.}
We jointly train CamAct on robot demonstrations and
camera-derived pseudo-actions using conditional flow matching.
With Gaussian noise $\boldsymbol{\epsilon}$ and flow time
$s\in[0,1]$, we construct
$\mathbf{Y}_s=(1-s)\boldsymbol{\epsilon}+s\mathbf{Y}$
and define the velocity prediction residual as
$\mathbf{E}_{\theta}
=\mathbf{V}_{\theta}(\mathbf{Y}_s,s,\mathbf{h}_t)
-(\mathbf{Y}-\boldsymbol{\epsilon})$.

We define source-specific masks
$\mathbf{m}^{R}=[\mathbf{1}_{d_R};\mathbf{0}_{3}]$
and
$\mathbf{m}^{V}=[\mathbf{0}_{d_R};\mathbf{1}_{3}]$,
and broadcast them across the prediction horizon to obtain
$\mathbf{M}^{q}$ for $q\in\{R,V\}$.
The masked loss is
\begin{equation}
\mathcal{L}_q =
\mathbb{E}_{s,\epsilon}
\left[
\frac{\left\|M_q \odot E_\theta\right\|_F^2}
{\max\left(1,\sum_{h,j} M_{q,h,j}\right)}
\right].
\label{eq:masked_loss}
\end{equation}
where the denominator counts the active supervision
entries across time steps and output dimensions.
This normalization prevents one source from receiving
greater loss weight simply because it supervises more
output dimensions.
The masks apply only to the loss.

We sample robot and video examples with probabilities
$1-\lambda$ and $\lambda$, respectively, and average
their per-sample losses, yielding the expected objective
\begin{equation}
    \mathcal{L}_{\mathrm{joint}}
    =
    (1-\lambda)
    \mathbb{E}_{\mathcal{D}_R}[\mathcal{L}_R]
    +
    \lambda
    \mathbb{E}_{\mathcal{D}_V}[\mathcal{L}_V].
    \label{eq:objective}
\end{equation}
We jointly optimize the shared backbone and action head
end-to-end on this mixture, without additional
source-specific loss weighting.

At inference, we extract and denormalize the robot action
coordinates from the generated chunk for execution.
Training and inference configurations are provided
in the supplementary material.

\section{Experiments}
\label{sec:experiments}

\subsection{Simulation Experiments}
\label{sec:sim_experiments}

\paragraph{Experimental Setup.}
We instantiate CamAct with a Qwen3-VL-4B backbone and
evaluate two variants using either the GR00T or $\pi$
flow-matching head.
We jointly train each variant on robot demonstrations
and camera-derived pseudo-actions from in-the-wild videos,
using a robot-to-video sampling ratio of 80:20 unless
otherwise specified.
The backbone and action head are optimized end-to-end.
We compare each variant with its corresponding robot-only
baseline trained for the same number of updates.

\paragraph{Benchmarks and Evaluation.}
We conduct simulation experiments on LIBERO and RoboCasa 365.
On LIBERO, we evaluate four task suites: Spatial, Object,
Goal, and Long.
On RoboCasa 365, we train on Human300 and evaluate on
Target50, which consists of 18 atomic seen, 16 composite
seen, and 16 composite unseen tasks.
We perform 50 evaluation rollouts per task on both
benchmarks and report success rates for individual suites
or task categories and the overall benchmark.
In addition, we assess out-of-distribution generalization
through zero-shot evaluation on LIBERO-Plus using policies
trained on LIBERO.

\paragraph{Main Results.}
As shown in Table~\ref{tab:manipulation_success}, CamAct
improves overall success rates on both benchmarks with
both action heads.
On LIBERO, both variants achieve an overall success rate
of 98.10\%, outperforming their corresponding robot-only
baselines by 1.70 percentage points with GR00T and
1.20 points with $\pi$.
The largest gains occur on LIBERO-Long, where success
rates increase from 91.20\% to 96.80\% with GR00T
and from 93.40\% to 96.20\% with $\pi$.

On RoboCasa 365, CamAct improves overall success from
22.96\% to 29.32\% with GR00T and from 25.16\% to
27.68\% with $\pi$, corresponding to gains of 6.36 and
2.52 percentage points, respectively.
The GR00T variant improves across all three task categories,
while both variants achieve higher success on unseen
composite tasks.

In addition, CamAct improves the success rate on LIBERO-Plus
from 71.90\% to 74.49\% (Table~\ref{tab:libero_plus}).

\begin{table}[t]
\centering
\caption{OOD evaluation on LIBERO-Plus.
We report success rates (\%) without training on LIBERO-Plus.}
\label{tab:libero_plus}
\small
\setlength{\tabcolsep}{6pt}
\renewcommand{\arraystretch}{1.1}

\begin{tabularx}{\columnwidth}{Xc}
\toprule
Method & Success Rate (\%) \\
\midrule
OpenVLA-OFT~\cite{kim2025oft} & 69.6 \\
$\pi_0$-FAST~\cite{pertsch2025fast} & 61.6 \\
$\pi_0$~\cite{black2024pi0} & 53.6 \\
GR00T N1.6~\cite{bjorck2025gr00t} & 72.6 \\
StarVLA-GR00T~\cite{community2026starvla} & 71.90 \\
\midrule
\rowcolor[gray]{0.93}
CamAct (Ours) & 74.49 \\
\bottomrule
\end{tabularx}
\end{table}

\subsection{Real-World Experiments}
\label{sec:real_experiments}

\begin{figure*}
    \centering
    \includegraphics[width=\linewidth]{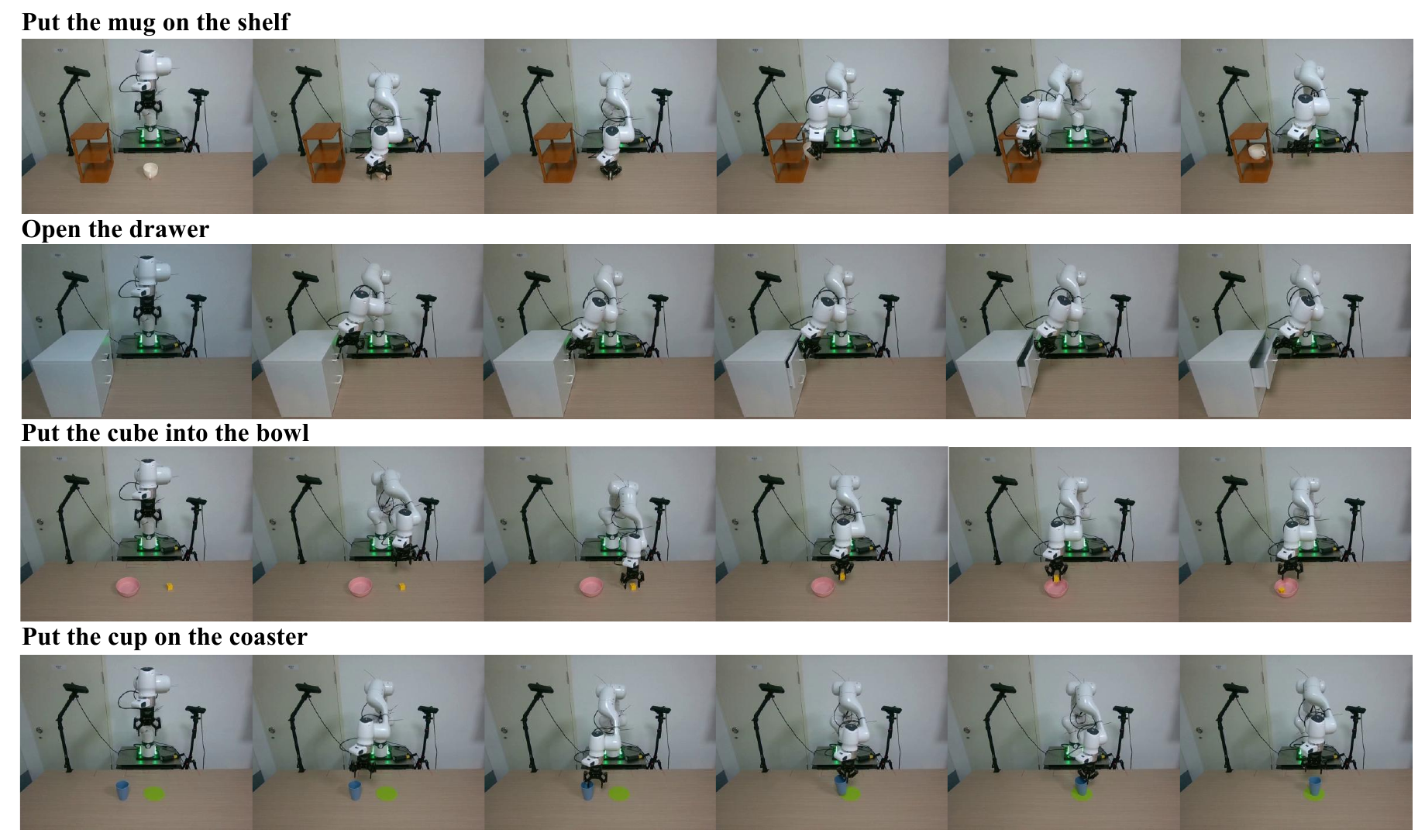}
    \caption{Visualization of the four real-world
    manipulation tasks.}
    \label{fig:real_tasks}
\end{figure*}

\paragraph{Experimental Setup.}
We evaluate CamAct on a Franka Research 3 robot equipped with an external camera and a wrist camera.
We consider four manipulation tasks
(Fig.~\ref{fig:real_tasks}): placing a cup on a shelf, opening a drawer, placing a block in a bowl, and placing a cup on a coaster. For each task, we collect 50 robot demonstrations and fine-tune a separate policy from a checkpoint trained on LIBERO, using the GR00T action head. We compare CamAct with its corresponding robot-only baseline over 16 trials per task, using matched initial configurations across the two methods. Each trial is limited to 300 steps. For the block-in-bowl and cup-on-coaster tasks, object positions are randomized during evaluation.

\begin{figure}
    \centering
    \includegraphics[width=\linewidth]{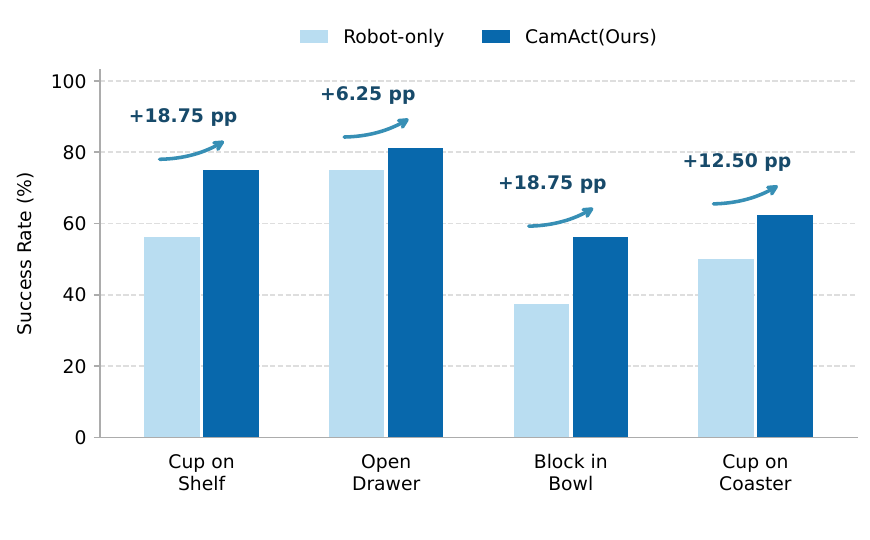}
    \caption{Real-world manipulation success rates.
    Each method is evaluated over 16 trials per task
    using matched initial configurations.}
    \label{fig:real_world}
\end{figure}

\paragraph{Results.}
As shown in Fig.~\ref{fig:real_world}, CamAct improves
the overall success rate from 54.69\% to 68.75\%,
a gain of 14.06 percentage points over the robot-only
baseline.
CamAct achieves more successful trials on all four tasks.
The largest gains occur on cup-on-shelf and block-in-bowl,
with three additional successful trials on each task,
corresponding to improvements of 18.75 percentage points.
On drawer-opening and cup-on-coaster, CamAct improves
success rates by 6.25 and 12.50 percentage points,
respectively.
The gains on block-in-bowl and cup-on-coaster are observed
under randomized initial object positions. These results suggest that the benefits of joint training with camera-motion supervision extend beyond simulation to real-world manipulation after task-specific fine-tuning.

\subsection{Ablation Studies}
\label{sec:ablation}

To understand the role of camera-motion supervision in
manipulation learning, we investigate the following questions:
(1) How does the choice of observation views affect the
benefit of camera-motion supervision?
(2) How do video observations contribute to learning
from camera-motion supervision?
(3) How do source video properties affect downstream
manipulation performance?
(4) How does the robot-to-video sampling ratio affect
performance?
Additional ablation results are provided in the
supplementary material.

\paragraph{Effect of Observation Views.}
We investigate how the observation views affect the benefit of camera-motion supervision. On LIBERO, we compare three observation configurations: using the third-person camera alone, the wrist camera alone,
or both cameras. For each configuration, we compare CamAct with a robot-only baseline under matched training settings.

As shown in Table~\ref{tab:wrist_ablation}, CamAct improves success rates over the robot-only baseline by 2.10 percentage points when using the wrist camera alone and by 1.70 percentage points when using both cameras. In contrast, success drops by 1.90 percentage points when using only the third-person camera. These results suggest that the benefit of camera-motion supervision depends on the observation views, with positive transfer observed in both configurations that include a wrist camera.

\begin{table}[t]
\centering
\caption{Success rates (\%) on LIBERO with different camera views. $\Delta$ denotes the change from the robot-only baseline to CamAct in percentage points.}
\label{tab:wrist_ablation}
\small
\setlength{\tabcolsep}{4pt}
\renewcommand{\arraystretch}{1.1}
\begin{tabular}{@{}lccc@{}}
\toprule
Camera views & Robot-only & CamAct & $\Delta$ (pp) \\
\midrule
Third-person only    & \textbf{91.20} & 89.30          & $-1.90$ \\
Wrist only           & 94.80          & \textbf{96.90} & $+2.10$ \\
Both & 96.40          & \textbf{98.10} & $+1.70$ \\
\bottomrule
\end{tabular}
\end{table}

\paragraph{Effect of Video Supervision.} We examine the contributions of video observations and camera-motion supervision by comparing full CamAct with two variants (Table~\ref{tab:supervision_ablation}). CamAct w/o Motion retains the same video inputs, normalized motion targets, and sampling settings, but sets the entire action-loss mask for video samples to zero. These samples still undergo forward computation but contribute no supervised loss, while robot action supervision remains unchanged. CamAct w/o Vision replaces video observations with blank images while retaining camera-motion supervision.

\begin{table}[t]
\centering
\caption{Ablation of modalities supervision on RoboCasa 365.
$\Delta$ denotes the change relative to full CamAct
in percentage points.}
\label{tab:supervision_ablation}
\small
\setlength{\tabcolsep}{6pt}
\renewcommand{\arraystretch}{1.1}

\begin{tabular}{@{}lcc@{}}
\toprule
Setting & Success Rate (\%) & $\Delta$ (pp) \\
\midrule
CamAct w/o Motion   & 23.64 & $-5.68$ \\
CamAct w/o Vision   & 25.00 & $-4.32$ \\
CamAct              & \textbf{29.32} & -- \\
\bottomrule
\end{tabular}
\end{table}

Full CamAct achieves a success rate of 29.32\%, compared with 23.64\% for w/o Motion and 25.00\% for w/o Vision. Removing video supervision reduces success by 5.68 percentage points, supporting the benefit of the camera-motion objective. Replacing video observations with blank images reduces success by 4.32 percentage points, suggesting that visual conditioning makes this objective more effective for downstream manipulation.

\paragraph{Effect of Video Properties.}
We further examine how the properties of source videos affect downstream manipulation performance. We consider four properties: scene openness, cumulative rotation, relative translation, and turning concentration. For each property, we construct two video subsets with contrasting characteristics while keeping the amount of video training data fixed.

\begin{figure}[t]
    \centering
    \includegraphics[width=\linewidth]{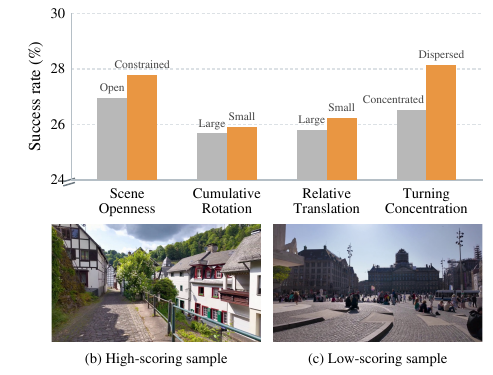}
    \caption{Effect of source video properties on manipulation performance. (a) Success rates using video subsets with contrasting properties and matched amounts of video training data. (b,c) Representative high-scoring and low-scoring samples, respectively.}
    \label{fig:video_properties}
\end{figure}

\begin{figure}[t]
    \centering
    \includegraphics[width=\linewidth]
    {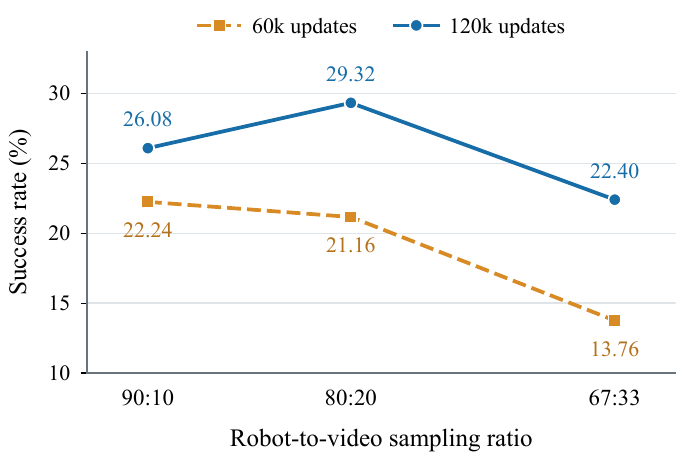}
    \caption{Effect of the robot-to-video sampling
    ratio on RoboCasa 365 success rates at different
    training budgets.}
    \label{fig:mixing_ratio}
\end{figure}

As shown in Fig.~\ref{fig:video_properties}(a), videos with more constrained scenes achieve slightly higher success rates than those with open scenes. Similarly, smaller cumulative rotations and relative translations yield modest improvements. The most pronounced difference appears in turning concentration, where videos with dispersed turning patterns outperform those with concentrated turns. Within the evaluated subsets, smaller camera motions and more dispersed turning patterns are associated with higher manipulation success.

\begin{figure*}[!t]
    \centering
    \includegraphics[width=\linewidth]{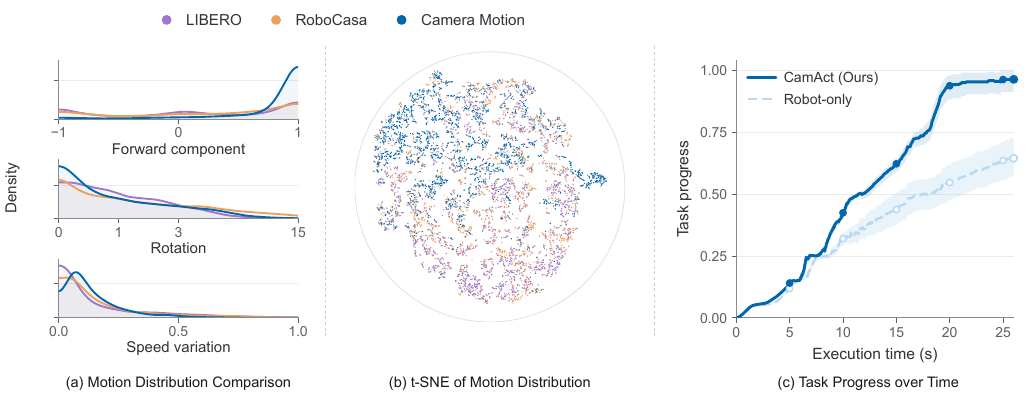}
    \caption{Motion distribution and task progress analysis.
    (a) Distributions of selected motion attributes
    for wrist-camera trajectories from LIBERO and
    RoboCasa 365 demonstrations and camera trajectories
    from in-the-wild videos.
    (b) Joint t-SNE visualization of motion features
    based on the planar motion representation used
    for video supervision during training.
    (c) Mean task progress on a LIBERO-10 task
    involving two moka pots.
    Shaded regions indicate pointwise 95\% bootstrap
    confidence intervals.}
    \label{fig:analysis}
\end{figure*}

\paragraph{Effect of Robot--Video Mixing Ratio.}
We study how the robot-to-video sampling ratio affects manipulation performance on RoboCasa 365. We compare three configurations---90:10, 80:20, and 67:33---at training budgets of 60k and 120k updates.

As shown in Fig.~\ref{fig:mixing_ratio}, the effect of the sampling ratio depends on the training budget. At 60k updates, 90:10 slightly outperforms 80:20, with success rates of 22.24\% and 21.16\%, respectively. At 120k updates, increasing the video sampling proportion from 10\% to 20\% improves success from 26.08\% to 29.32\%, while further increasing it to 33\% reduces success to 22.40\%. Based on the results at 120k updates, we adopt 80:20 as the default sampling ratio in our main experiments.

\subsection{Analysis}
\label{sec:analysis}

The ablations above show that the benefits of camera-motion supervision depend on observation views and training configurations. We next examine whether camera motion and robot wrist-camera motion exhibit similar statistical patterns. We also compare task progression between CamAct and the robot-only in a manipulation case study.

\paragraph{Motion Distributions.}
We reconstruct wrist-camera trajectories from LIBERO
and RoboCasa 365 demonstrations using end-effector poses, and compare them with camera
trajectories recovered from in-the-wild videos.
As shown in Fig.~\ref{fig:analysis}(a), video-camera
motion exhibits a stronger forward-direction bias,
while the three data sources show overlapping
distributions of the absolute up-axis rotation
component and speed variation.

We characterize motion windows using the same planar
motion components used for video supervision during
training: forward displacement, leftward displacement,
and heading change.
We jointly standardize the resulting motion features
from LIBERO, RoboCasa 365, and in-the-wild videos,
and visualize using t-SNE.
As shown in Fig.~\ref{fig:analysis}(b), video-camera
and robot wrist-camera motion also exhibit partially
overlapping distributions in the projection,
suggesting similar motion patterns under this representation.

\paragraph{Task Progress.}
We use a LIBERO-10 task involving two moka pots
as a case study to compare task progression between
CamAct and the robot-only baseline.
At each evaluation step, we compute progress based
on the states of both pots.
Each pot contributes four components---approaching,
grasping, transporting, and placing---giving eight
components combined into a normalized score
in $[0,1]$.
Each point in Fig.~\ref{fig:analysis}(c) represents
the mean progress over 50 rollouts at that time.
CamAct advances through the task more quickly
and achieves higher final mean progress than
the robot-only baseline. CamAct achieves both a higher task success rate
and shorter task completion times than
the robot-only baseline.

\section{Conclusion}
\label{sec:conclusion}

We presented CamAct, a unified VLA framework that jointly learns from robot demonstrations and camera motion from in-the-wild videos. CamAct converts camera trajectories into pseudo-actions and learns them alongside robot actions with masked supervision. Joint training improves success rates from 96.40\% to 98.10\% on LIBERO and from 22.96\% to 29.32\% on RoboCasa 365. CamAct improves out-of-distribution generalization on LIBERO-Plus and raises average success from 54.69\% to 68.75\% across four real-world manipulation tasks. Ablations and motion distribution analyses examine how video characteristics, data composition, and training design influence the benefits of camera-motion supervision. These findings show that in-the-wild videos can benefit robotic manipulation even when they contain no demonstrations of object manipulation.



{
     \small
     \bibliographystyle{ieeenat_fullname}
     \bibliography{main}
}


\clearpage
\appendix
\twocolumn[{%
  \noindent{\fontsize{25pt}{29pt}\selectfont\bfseries Appendix\par}%
  \vspace{8pt}%
}]

\section{Implementation Details}
\label{app:implementation_details}

\paragraph{Video Sources and Licensing.} Our video collection comprises YouTube videos that are explicitly labeled under the Creative Commons Attribution (CC BY) license. This license permits reuse and adaptation, including for research purposes, subject to appropriate attribution and the other applicable license conditions. We verify the license information of each video included in the collection and exclude videos marked under the Standard YouTube License or carrying additional usage restrictions. We use YouTube's official API in the collection process, following the applicable platform policies and license terms.

\paragraph{Video Preprocessing.} We sample the videos at 20\,Hz to match the control frequency of the simulation environments. We discard intervals where the relative camera pitch exceeds $35^\circ$, the planar speed exceeds $3\,\mathrm{m/s}$, or the yaw angular speed exceeds $60^\circ/\mathrm{s}$. We also remove near-static intervals lasting at least $1\,\mathrm{s}$ where both the planar speed is below $0.03\,\mathrm{m/s}$ and the yaw angular speed is below $2^\circ/\mathrm{s}$. We split trajectories at timestamp gaps exceeding $0.075\,\mathrm{s}$ and retain segments lasting at least $1.5\,\mathrm{s}$. After filtering, we smooth the planar positions and unwrapped headings using Gaussian-weighted local linear fitting with $\sigma=0.06\,\mathrm{s}$ and a temporal neighborhood of $\pm0.18\,\mathrm{s}$, retaining the segment endpoints. We then recompute the motion increments and check the speed and angular-speed limits again.

\subsection{Training Configuration}
We build CamAct on StarVLA using a Qwen3-VL-4B backbone and a flow-matching action head. In our main experiments, we use a robot-to-video sampling ratio of 80:20. Table~\ref{tab:training_details} summarizes the training hyperparameters.

\begin{table}[t]
    \centering
    \caption{Training hyperparameters for CamAct. Dataset-specific values are ordered as LIBERO / RoboCasa 365 / FR3.}
    \label{tab:training_details}
    \small
    \setlength{\tabcolsep}{4pt}
    \renewcommand{\arraystretch}{1.1}
    \begin{tabular}{@{}p{0.40\columnwidth}p{0.56\columnwidth}@{}}
        \toprule
        \textbf{Hyperparameter} & \textbf{Value} \\
        \midrule
        VLM backbone & Qwen3-VL-4B \\
        Action head & GR00T or $\pi$ (sim.); GR00T (real) \\
        Action horizon & 8 / 16 / 8 \\
        Optimizer & AdamW \\
        AdamW betas & $(0.9,\,0.95)$ \\
        AdamW epsilon & $10^{-8}$ \\
        VLM LR & $10^{-5}$ \\
        Action-head LR & $10^{-4}$ \\
        LR schedule & Warmup + cosine decay \\
        Warmup steps & 5,000 / 6,000 / 1,500 \\
        Weight decay & $10^{-8}$ \\
        Batch size & 128 / 128 / 64 \\
        Training updates & 30k / 120k / 30k \\
        Robot:video ratio & 80:20 \\
        \bottomrule
    \end{tabular}
\end{table}

\subsection{Inference Configuration}
After generating an action chunk, we extract and denormalize the robot-action coordinates for execution. The entire chunk is executed before the next prediction: 8 actions for LIBERO and real-world experiments, and 16 actions for RoboCasa 365. The control frequency is 20\,Hz in simulation and 15\,Hz in real-world experiments.

\begin{figure}[t]
    \centering
    \includegraphics[width=\linewidth]{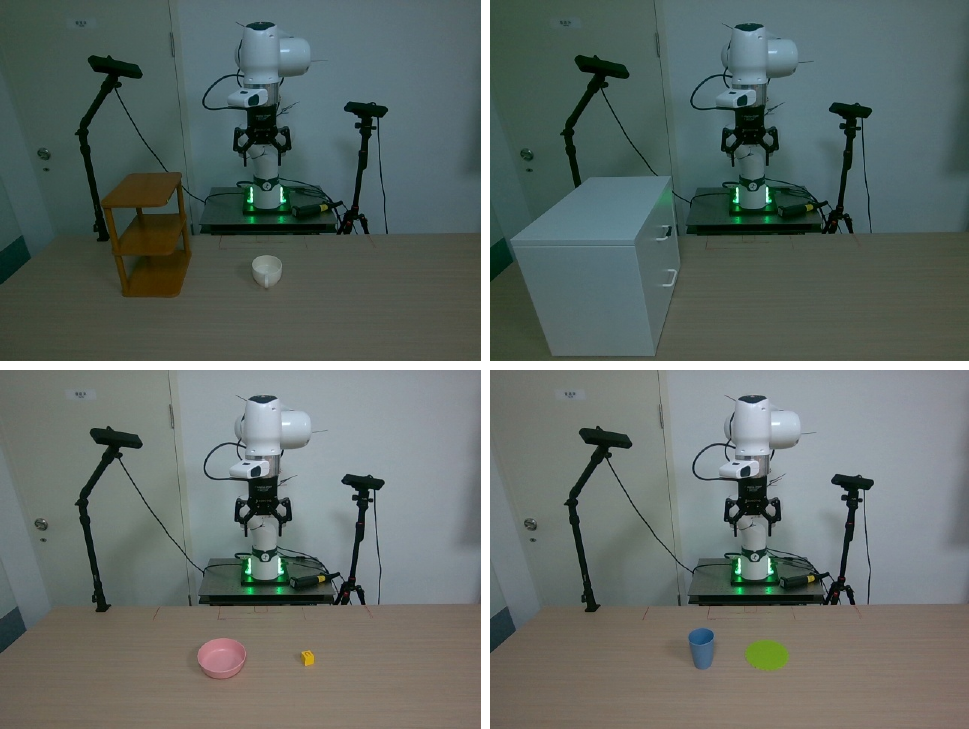}
    \caption{Initial scenes of the four real-world tasks, from left to right and top to bottom: \textit{Cup on Shelf}, \textit{Open Drawer}, \textit{Block in Bowl}, and \textit{Cup on Coaster}.}
    \label{fig:real_world_setup}
\end{figure}

\subsection{Real-World Experimental Setup}
We conduct real-world experiments on a Franka Research 3 robot equipped with an external camera and a wrist camera. We evaluate four tasks: \textit{Cup on Shelf}, \textit{Open Drawer}, \textit{Block in Bowl}, and \textit{Cup on Coaster}. For each task, we collect 50 demonstrations and fine-tune a separate policy. The robot-only baseline and CamAct are initialized from their respective LIBERO-trained checkpoints, with CamAct continuing to incorporate video supervision during fine-tuning.

Each method is evaluated over 16 trials per task using matched initial configurations, with a maximum of 300 steps per trial. Object positions are randomized for \textit{Block in Bowl} and \textit{Cup on Coaster}. Figure~\ref{fig:real_world_setup} shows the initial scenes of the four tasks.

\section{Additional Ablation Results}
\label{app:additional_ablations}

We provide additional ablations on LIBERO to examine the choice of action representation and the effect of freezing the vision-language backbone. Table~\ref{tab:additional_ablations} reports the results.

\begin{table}[t]
    \centering
    \caption{Additional ablation results on LIBERO. We report success rates (\%).}
    \label{tab:additional_ablations}
    \small
    \setlength{\tabcolsep}{5pt}
    \renewcommand{\arraystretch}{1.06}
    \begin{tabular}{@{}lc@{}}
        \toprule
        Setting & Success rate (\%) \\
        \midrule
        \multicolumn{2}{l}{\textit{Action representation}} \\
        Direct alignment & 95.75 \\
        Concatenation (CamAct) & \textbf{98.10} \\
        \midrule
        \addlinespace[2pt]
        \multicolumn{2}{l}{\textit{Frozen VLM}} \\
        Robot-only & 65.30 \\
        CamAct & \textbf{75.55} \\
        \bottomrule
    \end{tabular}
\end{table}

\paragraph{Action representation.} We compare concatenation with direct alignment. Direct alignment assigns the camera-motion components $(\Delta x,\Delta y,\Delta\psi)$ to zero-based action indices 0, 1, and 5, respectively. Concatenation instead preserves separate output coordinates for robot actions and camera-motion targets. It achieves 98.10\% success, compared with 95.75\% for direct alignment, a gain of 2.35 percentage points.

\paragraph{Frozen vision-language backbone.} We freeze the vision-language backbone and train only the action head for both the robot-only baseline and CamAct. Under this setting, CamAct improves success from 65.30\% to 75.55\%, a gain of 10.25 percentage points. These results show that camera-motion supervision remains beneficial even when the VLM is kept frozen.

\section{Additional Visualization}
\label{app:additional_visualization}

\begin{figure}[t]
    \centering
    \includegraphics[width=\linewidth]{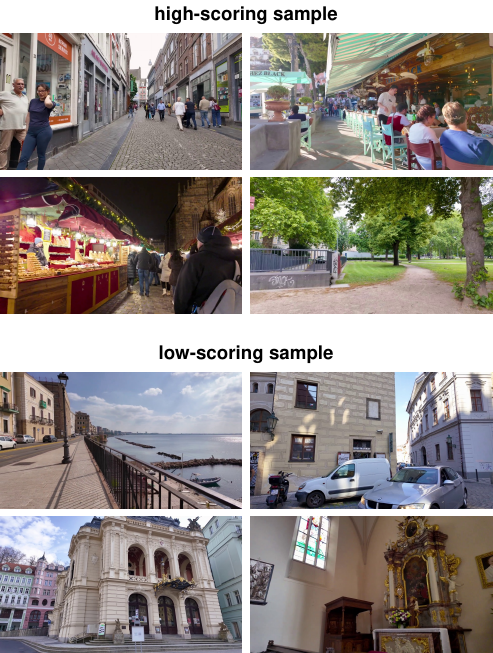}
    \caption{Additional high-scoring (top) and low-scoring (bottom) video samples.}
    \label{fig:additional_video_examples}
\end{figure}

\paragraph{High- and low-scoring video samples.} Figure~\ref{fig:additional_video_examples} provides additional high- and low-scoring video samples, complementing the examples in Fig.~5 of the main paper.

\begin{figure*}[t]
    \centering
    \includegraphics[width=\textwidth,height=0.92\textheight,keepaspectratio]{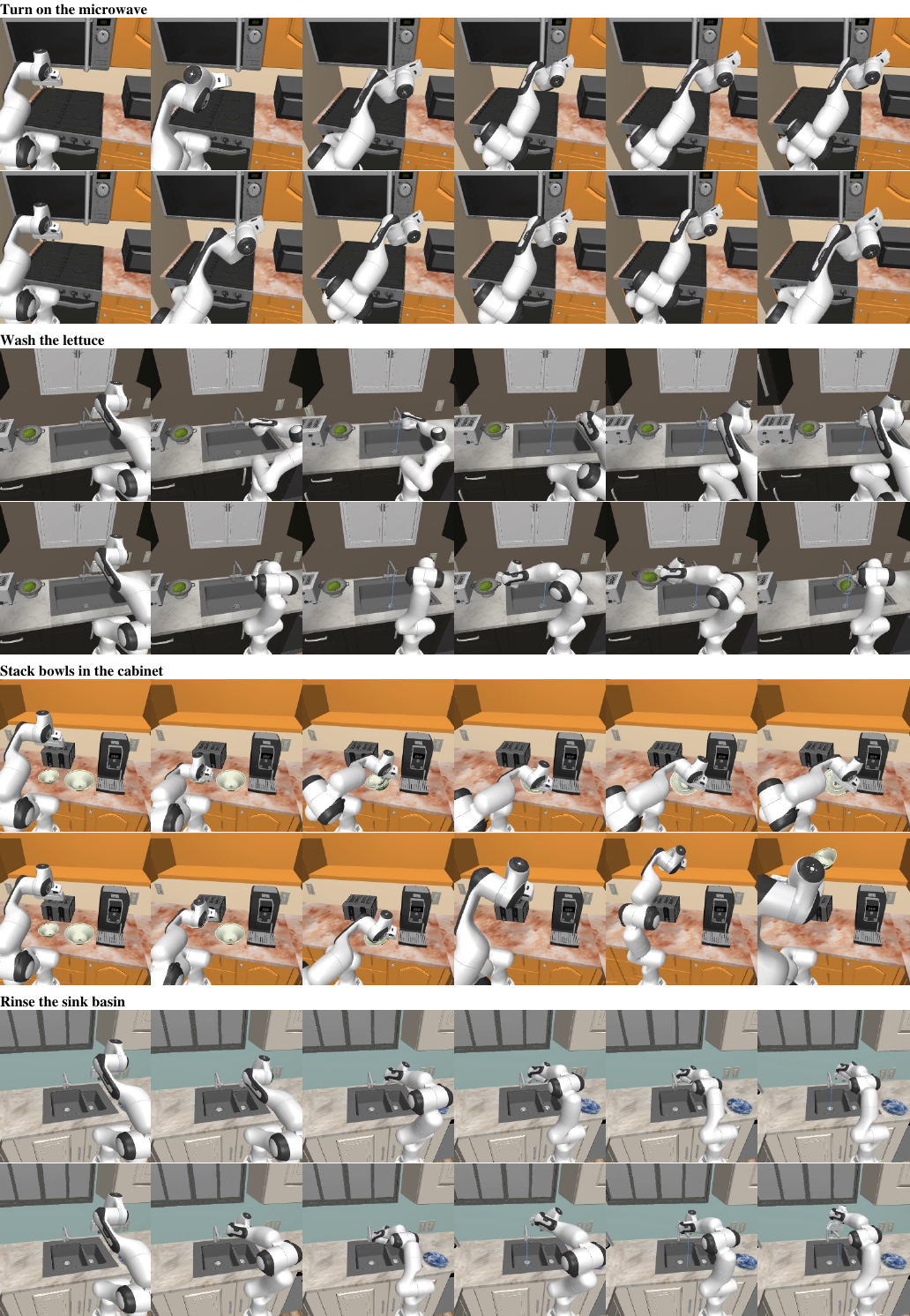}
    \caption{Qualitative comparison on four RoboCasa tasks. For each task, the top row shows the robot-only baseline and the bottom row shows CamAct.}
    \label{fig:simulation_comparison}
\end{figure*}

\newpage
\paragraph{Simulation rollouts.} Figure~\ref{fig:simulation_comparison} provides qualitative comparisons between CamAct and the robot-only baseline on four RoboCasa tasks using the checkpoints evaluated in the main paper. Each paired rollout starts from the same initial scene, and corresponding frames are sampled at the same simulation times. We select cases in which CamAct succeeds while the robot-only baseline fails within the evaluation horizon.

\end{document}